%% file: main.tex
\documentclass[10pt,twocolumn,letterpaper]{article}

\usepackage[pagenumbers]{cvpr} 

\input{preamble}

\definecolor{cvprblue}{rgb}{0.21,0.49,0.74}
\usepackage[pagebackref,breaklinks,colorlinks,allcolors=cvprblue]{hyperref}
\usepackage{arydshln}

\def\paperID{14458} 
\def\confName{CVPR}
\def\confYear{2025}

\title{IM-Portrait: Learning 3D-aware Video Diffusion for Photorealistic \\Talking Heads from Monocular Videos}

\author{
Yuan Li$^{1,2*}$\footnotemark[4] \quad
Ziqian Bai$^{2}$\footnotemark[1] \quad
Feitong Tan$^{2}$ \quad
Zhaopeng Cui$^{1}$\footnotemark[2] \quad
Sean Fanello$^{2}$ \quad \\
Yinda Zhang$^{2}$\footnotemark[2] \\
$^{1}$State Key Lab of CAD\&CG, Zhejiang University \quad
$^{2}$Google\\
}

\begin{document}

\twocolumn[{%
\renewcommand\twocolumn[1][]{#1}%
\maketitle
\begin{center}
    \centering
    \vspace{-2em}
    \captionsetup{type=figure}
    \includegraphics[width=\linewidth, trim={0 0 0 0}, clip]{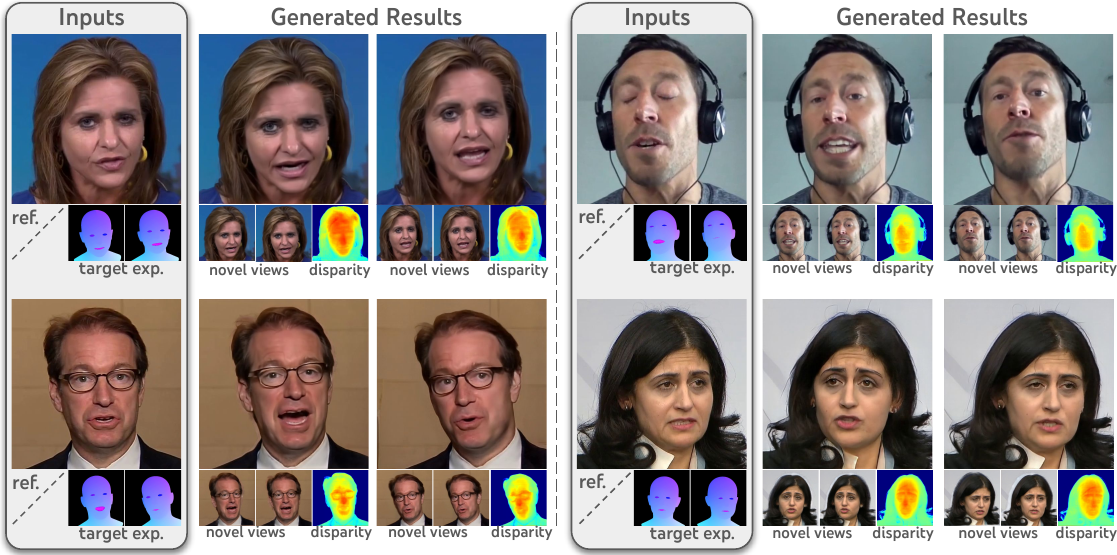}
    \vspace{-2em}
    \captionof{figure}{We propose a 3D-aware video diffusion model for talking head synthesis. Given an image as identity and a sequence of tracking signals (as shown on the left for each example), our model directly generates videos in Multiplane Images (MPIs) in a single denoising process, which is ready for efficient novel-view rendering. This enables immersive viewing experience, \eg rendering binocular stereo or perspective distortion in VR headset.
    Please see our website for more results: \href{https://y-u-a-n-l-i.github.io/projects/IM-Portrait/}{https://y-u-a-n-l-i.github.io/projects/IM-Portrait/}.}
    \label{fig:teaser}
\end{center}%
}]

\renewcommand{\thefootnote}{\fnsymbol{footnote}}
\footnotetext[1]{Authors contributed equally.}
\footnotetext[2]{Corresponding authors.}
\footnotetext[4]{Work was conducted while Yuan Li was an intern at Google.}

\input{sec/0_abstract}    
\input{sec/1_intro}
\input{sec/2_related_works}
\input{sec/3_method}
\input{sec/4_experiments}
\input{sec/5_conclusion}

{
    \small
    \bibliographystyle{ieeenat_fullname}
    \bibliography{main}
}

\input{sec/X_suppl}

\end{document}

%% file: preamble.tex
\usepackage{tabularx}
\usepackage{multicol}
\usepackage{multirow}
\usepackage{soul}
\usepackage{bm}
\usepackage{makecell}
\usepackage{colortbl}

%% file: sec/0_abstract.tex
\begin{abstract}
We propose a novel 3D-aware diffusion-based method for generating photorealistic talking head videos directly from a single identity image and explicit control signals (e.g., expressions). 
Our method generates Multiplane Images (MPIs) that ensure geometric consistency, making them ideal for immersive viewing experiences like binocular videos for VR headsets.
Unlike existing methods that often require a separate stage or joint optimization to reconstruct a 3D representation (such as NeRF or 3D Gaussians), our approach directly generates the final output through a single denoising process, eliminating the need for post-processing steps to render novel views efficiently.
To effectively learn from monocular videos, we introduce a training mechanism that reconstructs the output MPI randomly in either the target or the reference camera space. This approach enables the model to simultaneously learn sharp image details and underlying 3D information.
Extensive experiments demonstrate the effectiveness of our method, which achieves competitive avatar quality and novel-view rendering capabilities, even without explicit 3D reconstruction or high-quality multi-view training data.
\end{abstract}

%% file: sec/1_intro.tex
\section{Introduction}
\label{sec:intro}

Synthesizing realistic talking heads, has been a long-lasting challenging problem in computer vision for decades. The recent emergence of diffusion models within the generative AI domain offers a promising solution to this problem.
Significant advancements have been made in modeling talking heads using image or video diffusion models conditioned on tracking signals \cite{xie2024x,ma2024follow,ding2023diffusionrig,liu2024diffdub}, leveraging the remarkable generative capabilities of diffusion models.

Most diffusion-based approaches, however, operate in 2D image space, and the extension of these methods to 3D rendering remains largely unexplored. One potential approach involves applying 3D transformations to the tracking signal, such as parametric models or landmarks. 
However, this necessitates running the reverse diffusion process for each new camera view, leading to significant computational overhead. Alternatively, some diffusion-based approaches generate multi-view images, which often lack geometric consistency. 
While off-the-shelf methods like NeRF~\cite{deng2024portrait4dv2} or 3D Gaussians can be used to construct a 3D representation for efficient rendering, the inherent inconsistencies in the input multi-view images limit the quality of the final 3D reconstruction, leading to potential issues in multi-view and temporal consistency.

In this work, we present, for the first time, a diffusion based approach that produces 3D talking heads that can be rendered efficiently in varying camera poses.
We adopt Multiplane Image (MPI) as the 3D representation due to its remarkable rendering efficiency and quality, and the affinity to 2D images such that we can make good use of existing 2D diffusion models.
Essentially, we formulate the problem of synthesizing talking heads images as a denoising diffusion process, which is conditioned on a reference image (i.e. the subject identity) and a sequence of control signals from parametric models (i.e. expressions and head poses).
In a single denoising process, our model generates MPI videos that can be readily rendered in varying camera pose and support spatial viewing experience.

Our model is learned directly from a large scale monocular talking head videos in-the-wild.
Due to the missing of multi-view datasets, like previous works, the 3D consistency needs to be learned from the head pose variability between video frames.
To achieve this, it is required to construct MPIs in a camera that looks at the head from a different viewpoint than the target images.
Specifically, we use the viewpoint of the reference image to construct MPIs. The estimated target images can then be rendered from MPIs by moving the camera according to the relative head pose differences between the reference and target images.
In this way, however, the frontal views of MPIs will be misaligned with the input noisy images (in our case the target ground-truth images), which will break the training of the diffusion model.
To tackle the problem, we propose a novel training framework to bootstrap our model to synthesize pseudo noisy input images that align with the MPIs in referece cameras. 
With this, the MPIs are randomly generated from either the reference camera viewpoint or the target camera viewpoint during the training, which allows the diffusion model to learn both 3D information and fine details.

Our contributions are as follows. 
\begin{itemize}
    \item We propose a diffusion-based approach for 3D-aware talking heads generation.
    \item Our diffusion model directly generates 3D renderable talking heads in a single denoising process, eliminating the need for post-processing steps.
    \item We propose a novel training framework that enables our model to learn from monocular videos, and show how randomly selecting the camera viewpoint between reference and target images allows for better convergence and sharper results.
    \item Extensive experiments demonstrate that our method achieves comparable or superior image and video quality to existing diffusion-based 2D talking head methods, while also offering efficient and consistent multi-view rendering on par with specialized (i.e. non-diffusion based) 3D approaches.
\end{itemize}

%% file: sec/2_related_works.tex
\section{Related Works}

\noindent\textbf{2D Talking Heads.}
In recent year, there has been a great advancement on synthesizing 2D talking heads with neural networks. A recent line of works~\cite{wang2021one,drobyshev2024emoportraits,bounareli2023hyperreenact,doukas2021headgan,drobyshev2022megaportraits,yin2022styleheat,zakharov2020fast,siarohin2019first} propose to leverage
neural networks to encode the appearance and motion from input images~\cite{wang2021one,drobyshev2022megaportraits,drobyshev2024emoportraits,siarohin2019first} (or explicit control signals~\cite{doukas2021headgan,yin2022styleheat,zakharov2020fast}) to some latent features or manually defined representations (\eg key-points, warping fields, texture maps), which are injected into 2D CNN generators with various architectures for talking head synthesis.
More recently, many works~\cite{xie2024x,ma2024follow,ding2023diffusionrig,liu2024diffdub} started to adopt 2D diffusion models, leveraging their impressive image quality, for controllable talking head synthesis. However, these approaches usually use simplified camera models (\eg orthogonal projection) or they do not have any explicit camera control, making them unsuitable for spatial viewing experiences such as binocular stereo and perspective geometry.
In addition, these methods usually do not produce explicit 3D geometry, which also limits their downstream applications.
While most of prior work learns talking head models from monocular videos which are available in a large scale in-the-wild, multi-view datasets, \eg from lightstage \cite{guo2024liveportrait}, could be leveraged to further improve fidelity, however such data requires intensive collection labor and there are no large scale multi-view datasets that are publicly available.

\vspace{2mm}
\noindent\textbf{3D Talking Heads.}
The problem of generating 3D talking heads has been a hot topic in the community. People attempted to model their talking heads with various 3D representations, such as textured mesh~\cite{chaudhuri2020personalized,grassal2022neural}, Neural Radiance Fields (NeRFs)~\cite{mildenhall2021nerf} or other implicit functions~\cite{gafni2021dynamic,bai2023learning,bai2024efficient,athar2022rignerf,zielonka2023instant,gao2022reconstructing,deng2024portrait4d,deng2024portrait4dv2}, 
and, more recently, Gaussian Splatting~\cite{chen2024monogaussianavatar,saito2024relightable,qian2024gaussianavatars}. While most of these approaches are person-specific, which requires to retrain a new model for every new person, some methods~\cite{deng2024portrait4d,deng2024portrait4dv2} also explore the potential of person-agnostic models with explicit 3D representations. Despite having good 3D properties, these approaches cannot easily leverage the recent advancement on 2D diffusion, 
which lower the upper bound of their image quality.

\vspace{2mm}
\noindent\textbf{Diffusion-based 3D Generative Model.}
Another line of work, investigates how to apply diffusion models on 3D representations. The most straightforward way is applying the standard diffusion process directly on 3D data~\cite{wang2025mvdd,vahdat2022lion}. However, this requires large scale 3D datasets, which are generally not available. One typical workaround is to train the diffusion model with 2D images. Some works~\cite{lan2025ln3diff,dupont2022data,jun2023shap,muller2023diffrf,wang2023rodin} firstly learn a latent space of the 3D samples with multiview 2D images, 
then apply diffusion on the learned latent space. However, this results to an additional latent space learning, which is non-trivial thus increasing the pipeline complexity. Some other works~\cite{tewari2023diffusion,anciukevivcius2023renderdiffusion} directly model 3D scene distributions with image space diffusion by integrating the 3D-to-2D rendering into the reverse process. However, these methods are mostly proposed for static scenes where multiview datasets are available. In contrast to prior research, 
our method handles dynamic 3D talking heads and is trained exclusively on monocular RGB videos from real-world sources.

%% file: sec/3_method.tex
\section{Method}

Our goal is to generate 3D talking head videos represented as Multiplane Images (MPIs), conditioned on a reference image (the subject's identity) and a sequence of control signals from a parametric model to determine expressions and head poses. The generated videos must exhibit high image quality, controllability, and efficient rendering for VR scenarios, including binocular consistency.

Given the scarcity of large-scale multi-view video datasets, we design our training pipeline to operate solely on 2D monocular videos from the wild. To leverage the power of 2D diffusion models, we formulate our model as a video diffusion process operating in the image space. However, this raises two key challenges:
1) How to generate 3D scenes from a diffusion denoising process defined on 2D observations, without access to ground-truth MPIs? 2) How to learn 3D information from monocular 2D videos?

To address the first challenge, we design a video diffusion pipeline integrated with a differentiable forward rendering model~\cite{tewari2023diffusion} (\cref{sec:3d_diffusion_process}) to directly model 3D scene distributions from 2D monocular data. 

To address the second challenge, a possible approach involves constructing MPIs of a 3D
talking head in the reference image's camera, then moving the target camera to render the head in novel views during training. 
This allows us to aggregate multi-view observations of the head from different frames into a consistent 3D space to learn 3D shapes.  
To this end, we propose an alternating training strategy that switches between using the reference and target views as the frontal camera for MPI construction 
and compute the loss on target views accordingly (\cref{sec:training_framework}). This approach encourages the network to learn to synthesize high-quality, sharp images when building the MPIs in target views and to learn multiview consistency when building it in the reference view. When MPIs are constructed in the reference camera, 
the target ground-truth images cannot be used as input of the diffusion model, as they are not pixel-aligned with the MPIs. To tackle this, we bootstrap our model to generate the pseudo ground truth under the reference camera (\cref{sec:training_framework}).  During inference, our model generate 3D scene representations in a single denoising diffusion process without any post-processing or optimization (\cref{sec:inference}).

\begin{figure*}[htp]
    \centering
    \includegraphics[width=0.95\linewidth]{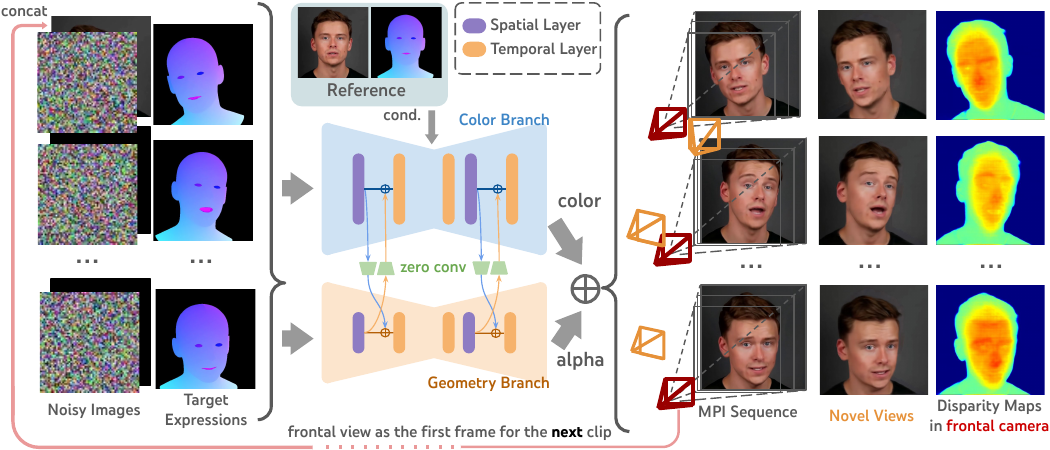}
    \vspace{-0.5em}
    \caption{\textbf{Inference pipeline.} Our model is built on the architecture of Lumiere~\cite{bar2024lumiere}, which takes an identity image, 2D noise video, the sequence of expressions rendered from 3DMM model and first frame image as input and outputs MPI video sequences. During inference, the network is conditioned on a reference portrait and takes the last frame of previously generated clip as the first frame condition. We separate the network into a color branch and a geometry branch. The two branches share information by zero convolution~\cite{zhang2023adding}. }
    \vspace{-1.5em}
\label{fig:pipeline}
\end{figure*}

\subsection{Preliminaries}

\noindent\textbf{Diffusion Formulation.}
Denoising diffusion probabilistic models (DDPM)~\cite{ho2020denoising} are one kind of generative models that learn to capture a data distribution $p_{\theta}(\mathbf{x})$. The method first constructs a forward Markovian process $q(\mathbf{x}_{0:T})$ to iteratively add noise to data as:
\begin{equation}\label{eq:forward_process}
    q(\mathbf{x}_t | \mathbf{x}_{t-1}) = \mathcal{N}(\mathbf{x}_t; \sqrt{1-\beta_t}\mathbf{x}_{t-1},\beta_t\mathbf{I}),
\end{equation}
where $\beta_t$ is scheduled to control the amount of noise being added to data in each step. A denoising diffusion model $g_{\theta}$ with parameters $\theta$ is trained to iteratively perform the reverse process $p_{\theta}(\mathbf{x}_{t-1}|\mathbf{x}_t)$ to generate a data sample $\mathbf{x}_0$ from a random Gaussian noise $\mathbf{x}_{T}$. If we parameterize the model $g_{\theta}$ to predict the estimated clean data sample $\hat{\mathbf{x}}_{t-1} = g_{\theta}(\mathbf{x}_t, t) \approx \mathbf{x}_0$, then the reverse process can be written as:
\begin{equation}\label{eq:reverse_process}
\begin{split}
    p_{\theta}(\mathbf{x}_{t-1}|\mathbf{x}_t) &= p_{\theta}(\mathbf{x}_{t-1}|\mathbf{x}_t, \hat{\mathbf{x}}_{t-1}) \\
    &= \mathcal{N}(\mathbf{x}_{t-1}; \bm{\mu}_t(\mathbf{x}_{t}, \hat{\mathbf{x}}_{t-1}), \sigma^{2}_{t}\mathbf{I})
\end{split}
\end{equation}
Please find the detailed formulation of $\bm{\mu}_t$ and $\sigma^{2}_{t}$ in Eq. (7) of DDPM~\cite{ho2020denoising}.

\vspace{2mm}
\noindent\textbf{Multiplane Images.}
\label{sec:mpi}
We leverage Multiplane Images (MPIs) to represent a 3D scene $\mathbf{S}$ (\eg a 3D talking head). An MPI representation consists of $D$ RGBA image planes placed in a reference 3D space defined by an MPI frontal camera $\phi^{\text{mpi}}$. These planes are parallel to the image plane of the frontal camera and equally spaced in disparity between near $d_{n}$ and far $d_{f}$ depths from the camera. During rendering, the planes are firstly warped from the frontal camera $\phi^{\text{mpi}}$ to a target camera $\phi^{\text{trgt}}$ by a homography. The warped planes are then composited via alpha blending. 
We also follow Tucker \etal~\cite{tucker2020single} to 
use only two RGB images, one as frontal, the other as residual, 
during MPI rendering to set a proper rendering capability of our model.

\subsection{3D Talking Head Video Diffusion}
\label{sec:3d_diffusion_process}

Unlike previous diffusion-based 2D talking head models ~\cite{xie2024x,ma2024follow}, our model aims to generate a sequence of 3D head representations in the form of an MPI video 
with $N$ frames $\{\mathbf{S}\}_{N}$. This is achieved by conditioning the model on a reference identity image $\mathbf{O}^{\text{ref}}$ and a sequence of control signals $\{\mathbf{C}\}_N$, which are the parameters of a mesh-based parametric head model~\cite{blanz2023morphable,li2017learning}, controlling facial expressions and head pose.
Inspired by recent 3D diffusion models~\cite{anciukevivcius2023renderdiffusion, tewari2023diffusion}, we propose a video diffusion framework that generates 3D content (an MPI video) directly from 2D observations by integrating the MPI rendering into the denoising process. More specifically, we formulate the forward process in our avatar diffusion as
\begin{equation}
    q(\mathbf{O}^{\text{mpi}}_{t} | \mathbf{O}^{\text{mpi}}_{t-1}) = \mathcal{N}(\mathbf{O}^{\text{mpi}}_{t};\sqrt{1-\beta_{t}}\mathbf{O}^{\text{mpi}}_{t-1}, \beta_t\mathbf{I}),
\end{equation}
where $\mathbf{O}^{\text{mpi}}$ is the 2D video observed by the MPI frontal camera $\phi^{\text{mpi}}$ defined in \cref{sec:mpi}. Note that we intentionally formulate the diffusion process on the image space of MPI frontal camera since it benefits the output quality. Also, we omit the $\{.\}_N$ in $\{\mathbf{O}^{\text{mpi}}\}_N$ for simplicity and do the same for $\{\phi^{\text{mpi}}\}_N$, $\{\mathbf{S}\}_N$ and $\{\mathbf{C}\}_N$. 
For the reverse process, instead of directly predicting the estimated clean 2D video $\hat{\mathbf{O}}^{\text{mpi}}_{t-1}$, we let the denoising model $g_{\theta}$ to predict the MPI video $\mathbf{S}_{t-1}$, followed by the MPI rendering to obtain $\hat{\mathbf{O}}^{\text{mpi}}_{t-1}$. Formally, we have
\begin{align}
\label{eq:forward_s}
    \mathbf{S}_{t-1} &= g_\theta(\mathbf{O}^{\text{ref}}, \mathbf{O}_{t}^{\text{mpi}}; t, \mathbf{C}^{\text{mpi}}) \\
\label{eq:forward_o}
    \hat{\mathbf{O}}_{t-1}^{\text{mpi}} &= \mathcal{R}(\mathbf{S}_{t-1}, \phi^{\text{mpi}}),
\end{align}
where $\mathcal{R}$ is the MPI rendering function.
Here, we rasterize the control signals $\mathbf{C}$ into UV coordinate maps under the MPI frontal camera $\phi^{\text{mpi}}$ to provide a pixel-aligned condition $\mathbf{C}^{\text{mpi}}$. We also convert the meshes of the parametric model under $\mathbf{C}$ into alpha planes and feed them into the model. In practice, we also concatenate the identity image $\mathbf{O}^{\text{ref}}$ with its corresponding UV coordinate map rasterized from the reference camera 
as the model input. Finally, we plug in the estimated clean 2D video $\hat{\mathbf{O}}^{\text{mpi}}_{t-1}$ to \cref{eq:reverse_process} to perform the reverse process, which is repeated $T$ times to sample the final MPI video output $\mathbf{S}_0$. Given a novel side view camera (or trajectory) $\phi^{\text{side}}$, we can simply run the MPI rendering to obtain the desired video $\mathbf{O}^{\text{side}}_0 = \mathcal{R}(\mathbf{S}_0, \phi^{\text{side}})$.

\subsection{Reference-Target Alternating Training}
\label{sec:training_framework}

\begin{figure*}[htp]
    \centering
    \includegraphics[width=0.95\linewidth]{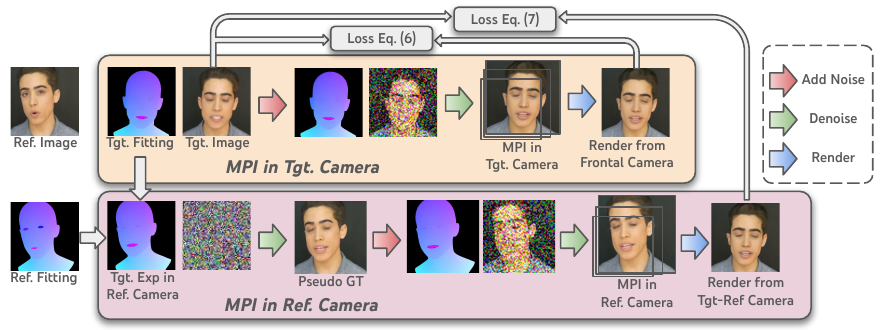}
    \vspace{-1em}
    \caption{\textbf{Illustration of Reference-Target Alternating Training.} On the top branch, we construct MPIs in the target camera and render the frontal view, where we directly use the ground truth target image to compute loss \cref{eq:main_loss_0} to learn sharp renderings and bootstrap the model. On the bottom branch, we construct MPIs in the reference camera. We first rasterize the target control signal (expression) in the reference camera, then use the bootstrapped model to generate pseudo ground truth by rendering the frontal view. The pseudo ground truth is added with noise, serving as the model input to compute loss \cref{eq:main_loss_1} for learning 3D shapes.
    }
    \vspace{-1.8em}
\label{fig:framwork}
\end{figure*}

Following a similar analysis as Tewari \etal~\cite{tewari2023diffusion}, a straightforward way to train our model in \cref{sec:3d_diffusion_process} needs two losses on videos rendered from two different cameras $\phi^{\text{mpi}}$ (the MPI frontal camera) and $\phi^{\text{side}}$ (a novel side view camera):
\begin{align}
    \hat{\mathbf{O}}_{t-1}^{\text{side}} &= \mathcal{R}(\mathbf{S}_{t-1}, \phi^{\text{side}}) \nonumber \\
\label{eq:main_loss_0}
    \mathcal{L}^{\text{mpi}}_{\theta} &= \mathbb{E}_{\mathbf{O}^{\text{ref}}, \mathbf{O}^{\text{mpi}}, \mathbf{C}^{\text{mpi}}, t}[\|\mathbf{O}^{\text{mpi}} - \hat{\mathbf{O}}^{\text{mpi}}_{t-1}\|^2] \\
\label{eq:main_loss_1}
    \mathcal{L}^{\text{side}}_{\theta} &= \mathbb{E}_{\mathbf{O}^{\text{ref}}, \mathbf{O}^{\text{mpi}}, \mathbf{C}^{\text{mpi}}, t, \mathbf{O}^{\text{side}}, \phi^{\text{side}}}[\|\mathbf{O}^{\text{side}} - \hat{\mathbf{O}}^{\text{side}}_{t-1}\|^2].
\end{align}
Intuitively, the first loss $\mathcal{L}^{\text{mpi}}_{\theta}$ encourages the model to give good frontal views of the MPIs, while the second loss $\mathcal{L}^{\text{side}}_{\theta}$ encourages the model to render good side views from the MPIs, which would be the only supervision for learning 3D shapes.
However, such approach would require a large scale multiview video dataset, which is not generally available.

In order to enable multi-view supervision when training purely on monocular 2D videos, a possible approach is to place MPIs in a canonical space with a fixed head pose, then move the rendering camera to simulate head pose changes.
In this way, the image data representing different head poses will act as a proxy ground truth for novel camera views to provide multi-view supervision. Therefore, we could separately compute loss $\mathcal{L}^{\text{mpi}}_{\theta}$ \cref{eq:main_loss_0} and loss $\mathcal{L}^{\text{side}}_{\theta}$ \cref{eq:main_loss_1} with only monocular 2D videos by using two different MPI setups as shown in \cref{fig:framwork}.
More specifically, when computing the first loss $\mathcal{L}^{\text{mpi}}_{\theta}$, we would set the MPI frontal cameras $\phi^{\text{mpi}}$ as the target image camera and use the target image as the frontal view ground truth $\mathbf{O}^{\text{mpi}}$.
Similarly, when computing the second loss $\mathcal{L}^{\text{side}}_{\theta}$, we would set the MPI frontal cameras $\phi^{\text{mpi}}$ to the reference image camera
and use the target image as the side view ground truth $\mathbf{O}^{\text{side}}$.

However, the above strategy only satisfies the ground truth requirements of the two losses, but does not consider the model input.  Indeed, when computing the second loss $\mathcal{L}^{\text{side}}_{\theta}$ \cref{eq:main_loss_1}, where the MPI frontal cameras $\phi^{\text{mpi}}$ are set to the reference camera, the training video frames are used as the ground truth to supervise the side view of MPIs. However these frames cannot be used as input of the diffusion model, as they are no longer pixel-aligned with the MPIs, which are constructed using the reference camera in this setting. 
To tackle this problem, we propose a bootstrapping training strategy that allows our model to generate the pseudo ground truth under the reference camera by itself.

\vspace{1mm}
\noindent\textbf{Bootstrapping.}
In order to bootstrap our model for generating pseudo ground truth under the reference camera, we firstly train our model $g_\theta(.)$ with only the loss $\mathcal{L}^{\text{mpi}}_{\theta}$ \cref{eq:main_loss_0} (\ie setting the MPI frontal cameras $\phi^{\text{mpi}}$ to the target camera). In this way, the model learns to generate plausible frontal views rendered from MPIs.

With this bootstrapping, we can then enable the second loss $\mathcal{L}^{\text{side}}_{\theta}$ \cref{eq:main_loss_1} to learn 3D structures of the MPIs. More specifically, we first set the MPI frontal cameras $\phi^{\text{mpi}}$ to the reference camera, then run the diffusion sampling with $k$ steps to generate the output MPI video $\mathbf{S}_k$, and render the frontal view to get images $\hat{\mathbf{O}}_{k}^{\text{mpi}}$.
Since the model has been trained to give plausible frontal views, the rendered images $\hat{\mathbf{O}}_{k}^{\text{mpi}}$ will have a reasonable quality, acting as the pseudo ground truth under the reference camera. Thus we can add noise to $\hat{\mathbf{O}}_{k}^{\text{mpi}}$ to obtain the model input $\mathbf{O}_{t}^{\text{mpi}}$ when computing the second loss $\mathcal{L}^{\text{side}}_{\theta}$ \cref{eq:main_loss_1} to train our model for learning 3D shapes inside MPIs.

\vspace{1mm}
\noindent\textbf{Late-stage Noise Sampling.}
When training with the second loss $\mathcal{L}^{\text{side}}_{\theta}$ \cref{eq:main_loss_1}, we compute the model input from the pseudo ground truth $\hat{\mathbf{O}}_{k}^{\text{mpi}}$, which cannot perfectly match the real image distribution. Such distribution mismatch could harm the model learning and subsequently lead to blurry results. To mitigate this negative influence, when training with loss $\mathcal{L}^{\text{side}}_{\theta}$ \cref{eq:main_loss_1}, we only sample $t$ from the second half of the forward Markovian process $[T/2, T]$ so that the added noise is relatively large thus has a better chance to hide the distribution mismatch.




\input{tabs/sota}

\vspace{1mm}
\noindent\textbf{Loss Functions.}
During training, we composite the MPIs and a background image calculated from the reference portrait $\mathbf{O}^{\text{ref}}$ with alpha blending to get the foreground renderings and the full image renderings. Using pre-computed foreground masks of the data videos, we calculate the standard L2 losses \cref{eq:main_loss_0} and \cref{eq:main_loss_1} on both the foreground renderings and the full image renderings.
Following Tewari \etal~\cite{tewari2023diffusion}, we additionally adopt several regularization losses besides the standard L2 losses \cref{eq:main_loss_0} and \cref{eq:main_loss_1}. The regularization losses include: 1) A LPIPS perceptual loss $\mathcal{L}_{\text{lpips}}$ for foreground renderings; 2) A mask loss $\mathcal{L}_{\text{mask}}$ to supervise the rendered alpha mask; 3) An edge-aware depth smoothing loss and a disparity loss following Tucker \etal~\cite{tucker2020single} to further supervise the underlying geometry, where we use the disparity maps rendered from the parametric model as pesudo ground truth for disparity loss.

\subsection{Run-time Inference}
\label{sec:inference}

When generating a 3D talking head from a driving video, we set the MPI frontal cameras $\phi^{\text{mpi}}$ to the driving video.
In order to generate long video sequence and smooth the discontinuity between adjacent video clips, we 
input ground truth first frame to the model in a similar way as Lumiere~\cite{bar2024lumiere} with random drop during training.
During inference, we take the last frame from the previous clip as the first frame of next clip. 
The generated first frame is then dropped due to overlapping when concatenating two adjacent clips.


%% file: tabs/sota.tex
\begin{table*}
  \centering
  \small
    \begin{tabular}{@{}c cccc cccc c}
        \toprule
        \multirow{2}{*}{Method} & \multicolumn{4}{c}{HDTF} & \multicolumn{4}{c}{Talkinghead1kh} &
        \multirow{2}{*}{\makecell{ Novel View \\  Render Speed}} \\ 
        \cline{2-5} \cline{6-9}
        & L1 $\downarrow$ & LPIPS $\downarrow$ & FID $\downarrow$ & FVD $\downarrow$ & L1 $\downarrow$ & LPIPS $\downarrow$ & FID $\downarrow$ & FVD $\downarrow$ \\ \hline
        Face-V2V~\cite{wang2021one} & \cellcolor[HTML]{FFA0A0}0.029 & 0.132 & 20.57 & \cellcolor[HTML]{FFD0D0}78.11 & \cellcolor[HTML]{FFA0A0}0.040 & 0.186 & 30.52 & \cellcolor[HTML]{FFD0D0}141.02 & 22 {\small FPS} in $256^2$ \\ 
        EmoPortrait~\cite{drobyshev2024emoportraits} & 0.039 & 0.189 & 24.77 & 140.51 & 0.058 & 0.252 & 41.08 & 280.22 & 7 {\small FPS} in $512^2$ \\
        \hline
        Portrait4d-v2~\cite{deng2024portrait4dv2} & - & - & 25.74 & 139.11 & - & - & 37.38 & 227.32 & 10 {\small FPS} in $512^2$\\
        \hline
        Follow-your-emoji~\cite{ma2024follow} & 0.035 & \cellcolor[HTML]{FFA0A0}0.115 & \cellcolor[HTML]{FDEDEC}16.09 & \cellcolor[HTML]{FDEDEC}117.54 & 0.047 & \cellcolor[HTML]{FFA0A0}0.159 & \cellcolor[HTML]{FFA0A0}19.37 & 209.91 & 0.5FPS* in $512^2$ \\
        X-Portrait~\cite{xie2024x} & \cellcolor[HTML]{FDEDEC}0.034 & \cellcolor[HTML]{FDEDEC}0.119 & \cellcolor[HTML]{FFA0A0}15.66 & 132.90 & \cellcolor[HTML]{FFD0D0}0.045 & \cellcolor[HTML]{FFD0D0}0.164 & \cellcolor[HTML]{FFD0D0}19.86 & \cellcolor[HTML]{FDEDEC}183.05 & 0.1FPS* in $512^2$ \\
        \hline
        Ours & \cellcolor[HTML]{FFD0D0}0.031 & \cellcolor[HTML]{FFD0D0}0.118 & \cellcolor[HTML]{FFD0D0}15.93 & \cellcolor[HTML]{FFA0A0}69.04 & \cellcolor[HTML]{FFD0D0}0.045 & \cellcolor[HTML]{FDEDEC}0.180 & \cellcolor[HTML]{FDEDEC}27.83 & \cellcolor[HTML]{FFA0A0}135.12 & 109 {\small FPS} in $512^2$\\
        \hdashline
        Ours w/o residual & 0.030 & 0.116 & 14.76 & 65.75 & 0.044 & 0.174 & 28.19 & 127.11 & 109 {\small FPS} in $512^2$\\
        Ours w/o late-stage & 0.031 & 0.123 & 16.70 & 70.69 & 0.045 & 0.180 & 29.86 & 136.07 & 109 {\small FPS} in $512^2$\\
        \hline
        \bottomrule
    \end{tabular}
    \vspace{-1em}
    \caption{\textbf{Quantitative comparison and ablation study.} We compare our method with previous work on both HDTF and Talkinghead1kh datasets. Our method achieves comparable image quality (L1, LPIPS, FID) and the best video quality (FVD) and novel-view rendering efficiency overall. Compared to ablation cases, the averaged metrics are close, and we highlight the actual huge gaps in perceived image, geometry, and side view rendering quality in \cref{fig:ablation_on_special_noise} and \cref{fig:ablation_on_residual_rgb}. All the render speeds are tested on a single Nvidia A100 GPU. *We report the generation speeds of diffusion models for Follow-your-emoji~\cite{ma2024follow} and X-Portrait~\cite{xie2024x}, as they do not support explicit camera control.}
    \vspace{-2.0em}
  \label{tab:self-reenact}
\end{table*}

%% file: sec/4_experiments.tex
\begin{figure*}[htp]
    \centering
    \includegraphics[width=\linewidth]{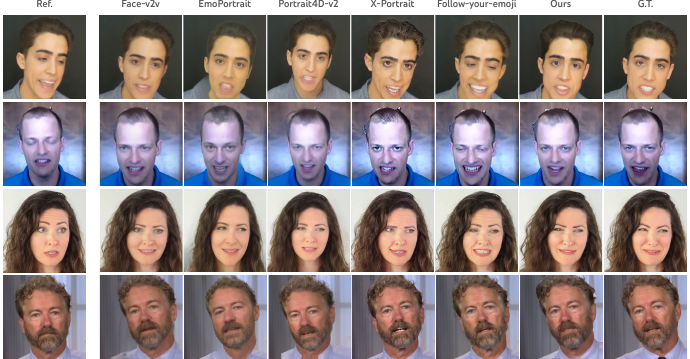}
    \vspace{-2em}
    \caption{\textbf{Talking head results.} We show results from previous work and our method. Our method generates results with sharp appearance and closely aligned expression with the ground truth. Note Portrait4d-v2 uses a customized camera space which is non-trivial to render GT camera aligned images.}
    \vspace{-1.7em}
\label{fig:comparison}
\end{figure*}

\section{Experiments}
In this section, we evaluate our method by comparing with prior arts on talking heads image quality, temporal quality, multi-view consistency, and computational efficiency. We also conduct qualitative comparisons and ablation studies.

\vspace{1mm}
\noindent\textbf{Implementation Details.} We train our model on a self-collected dataset consisting of 35k in-the-wild talking head monocular videos. Each video has a length between 4 to 6 seconds, and we crop the face region in $512 \times 512$ for training. Our model generates MPI video with 32 frames in a single network forwarding, and each MPI consists of 24 planes. We dynamically set the near and far planes based on the distance between the 3DMM head joint and the camera position. 
We optimize our model end-to-end on 8 NVIDIA H100 GPUs with a batch size of 8. During inference, we use DDIM sampling and do classifier-free guidance on first-frame and reference portrait with two scales~\cite{brooks2023instructpix2pix}. The guidance scale for the reference portrait is set to 1.5, while the guidance scale for the first frame is linearly reduced from 1 to 0 between frames 1 and 16. For the remaining 16 frames, the guidance scale for the first frame is fixed at 0.

\vspace{1mm}
\noindent\textbf{Benchmark.} We perform our qualitative comparison on two widely used public talking-head datasets, Talkinghead1kh~\cite{wang2021one} and HDTF~\cite{zhang2021flow} datasets. For HDTF dataset, we randomly sample video clips from 100 identities and each video clip consists of 180 frames. For Talkinghead1kh dataset, we randomly sample two 180-frame video clip from all evaluation identities. 
During testing, the first frame of each video clip is chosen as reference portrait $\mathbf{O}^{\text{src}}$ and all the animations are performed in a self-reenactment way. Cross-identity driving results are provided in the supp.

\vspace{1mm}
\noindent\textbf{Baselines.} We compare our method with existing one-shot photoreal talking heads approaches trained from monocular videos in the wild.
Specifically, we compare with representatives for three families of approaches: 1) Feed-forward models (\eg Face-V2V~\cite{wang2021one} and EmoPortrait~\cite{drobyshev2024emoportraits}). These methods do not use diffusion model but directly build 3D implicit feature volumes (as appearance) driven by warping fields (as expression). These methods are often optimized for the controllability in 2D space, but not the generation quality or 3D consistency. 
2) 3D NeRF-based one-shot talking heads (\eg Portrait4D-v2~\cite{deng2024portrait4d}), which establish the standard for 3D consistency due to the use of native 3D representation. 3) Video diffusion based methods (\eg follow-your-emoji ~\cite{ma2024follow} and xportrait~\cite{xie2024x}), which tend to be particularly strong on the quality of the generated images or videos. We perform all the evaluations on 512x512 resolution since most of the baselines outputs results at it, except Face-V2V for which we upsample its outputs to the desired resolution.

\subsection{Appearance and Motion Quality}

Following prior works, we compare image quality with L1, LPIPS~\cite{zhang2018unreasonable} and FID score~\cite{heusel2017gans} and video quality, a combination of images and motions, using FVD score~\cite{unterthiner2018towards}.  Note that L1 and LPIPS are not available for Portrait4D-v2 due to misalignments between its outputs and ground truth.

Quantitative results are shown in \cref{tab:self-reenact}. While the ranking based on different metrics varies, our method remains competitive among the state-of-the-art. Particularly, our model consistently achieves the best FVD on both datasets indicating our overall best video quality, taking into consideration both image quality and motion naturalness. This also indicates that a native video diffusion model is stronger than a finetuned image diffusion with added temporal component (like 
X-Portrait \cite{xie2024x} and Follow-your-emoji \cite{ma2024follow}).

We also show qualitative results in \cref{fig:comparison}.  Face-V2V~\cite{wang2021one} gives stable results and follows the input identity, but produces less details due to its low output resolution. EmoPortrait~\cite{drobyshev2024emoportraits} gives sharper results, but suffers from larger identity shifts. Portrait4D-v2~\cite{deng2024portrait4dv2} produces good sharpness and less identity shifts, however relatively stiff expressions. The diffusion-based methods Follow-your-emoji ~\cite{ma2024follow} and X-portrait~\cite{xie2024x} render good high-frequency details, but do not precisely follow the input head poses or expressions. 
In contrast, our method gives good image quality, faithful identity, while follows the input control signal.

\begin{figure*}[thp]
    \centering
    \includegraphics[width=1.0\linewidth]{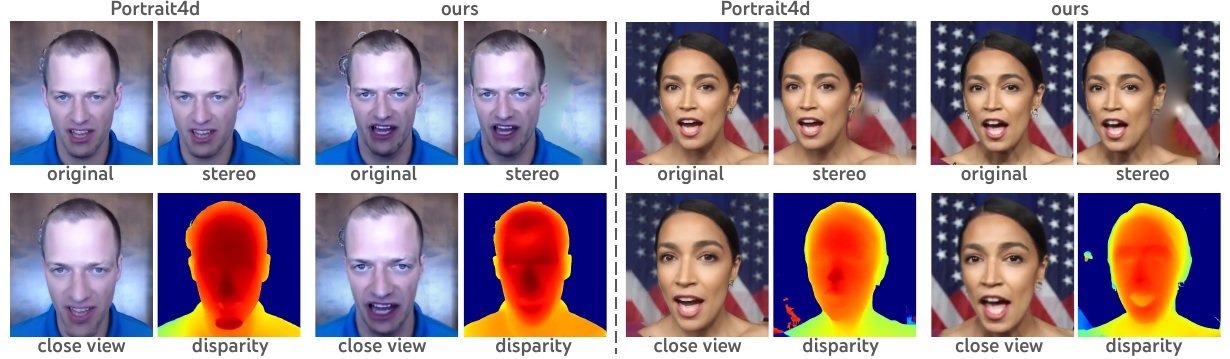}
    \vspace{-2em}
    \caption{\textbf{Binocular stereo and perspective effects.} Our method generates MPI videos that enables efficient spatial rendering, \eg binocular stereo and perspective effects. We render stereo pairs and 
    calculate a disparity using RAFT-Stereo~\cite{lipson2021raft} to visualize the perceived geometry. To demonstrate perspective effects, we render a view with camera moving closer to the subject.
    Our method renders visually plausible and comparable 3D effects with NeRF based approach like Portrait4d-v2, while being orders of magnitude faster.
    }
    \vspace{-1.5em}
\label{fig:pespective_effect}
\end{figure*}

\subsection{Effectiveness for 3D Rendering}
\label{sec:effect_on_3d}
We evaluate the effectiveness on 3D of different methods from the following aspects. First, we consider the task of binocular stereo rendering, which is one of the foundational use cases for VR/AR applications. Also, we compare how well these approaches perform on perspective geometry by moving the camera closer.

Note that Face-V2V~\cite{wang2021one} and EmoPortrait~\cite{drobyshev2024emoportraits} use orthogonal projection as a simplified camera model, thus they cannot produce any reasonable renderings for stereo pairs (\ie two views will have the same disparity across the whole image) and moving camera closer (\ie getting identical renderings). Follow-your-emoji ~\cite{ma2024follow} and X-xportrait~\cite{xie2024x} only use detected landmarks and images as the driving signal but do not allow any explicit control on the camera, 
thus they cannot perform these two 3D tasks neither.
As a result, we only show the renderings from Portrait4D-v2~\cite{deng2024portrait4dv2} and ours, where truth 3D rendering is supported.

\noindent\textbf{Binocular Stereo Rendering.}
We render stereo pairs with a $5$cm baseline looking from $1$m in front of the talking head, and run an off-the-shell stereo matching method~\cite{lipson2021raft} to get the disparity map. In \cref{fig:pespective_effect}, ours gives stereo pairs with reasonable estimated disparities, and is competitive with the NeRF-based Portrait4D-v2~\cite{deng2024portrait4dv2}, which acts as a strong baseline on 3D effects, while being one order of magnitude faster in novel view rendering (\cref{tab:self-reenact}).

\noindent\textbf{Perspective Geometry.}
To demonstrate the perspective effect, we render a close view by moving the camera closer while reducing camera focal length to keep the whole head in view. As in \cref{fig:pespective_effect}, ours produces good perspective distortions that is comparable to the NeRF-based Portrait4D-v2~\cite{deng2024portrait4dv2}, while being more faithful to the reference identity shown in \cref{fig:comparison} second row.

\subsection{Ablation}

\begin{figure}[t]
    \centering
    \includegraphics[width=1.0\linewidth]{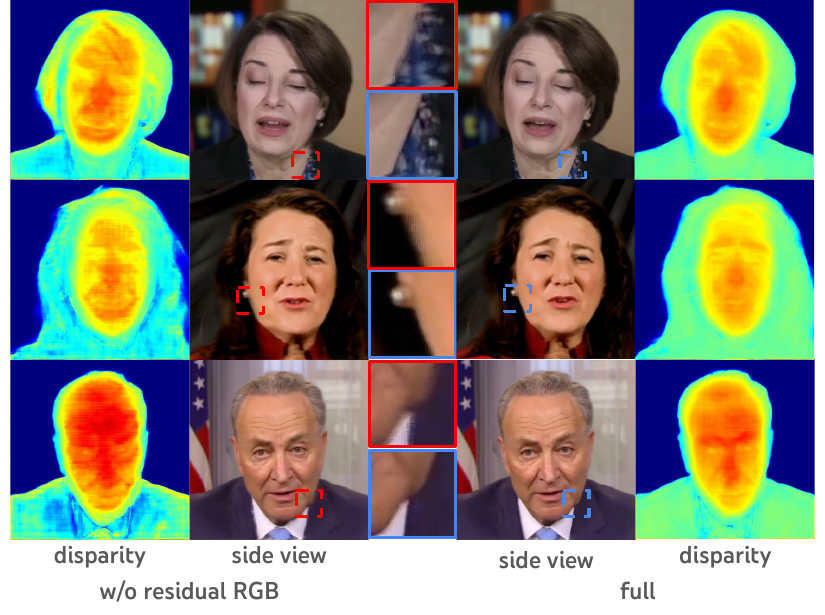}
    \vspace{-2em}
    \caption{\textbf{Effect of residual RGB.} We retrain our model with only a single frontal image in the MPI. We show both the rendered side view images and the disparity converted from the MPI alpha in frontal view. We found this significantly reduces the capacity of MPI, which leads to noisy and irregular geometry and blatant blurriness near the occlusion boundaries.}
    \vspace{-1.5em}
\label{fig:ablation_on_residual_rgb}
\end{figure}

\noindent\textbf{MPI as representation.}
Our diffusion model generates MPI frames, where each of them has a frontal, a residual, and a set of planes with alpha.
While the frontal contains most of the appearance, we hereby evaluate the importance of the residual image.
We train our model producing a MPI with only a frontal image, and reuse it for all the planes for alpha composition.
While the metrics on frontal view become even stronger compared to the full model (\ref{tab:self-reenact} (Ours w/o residual)), the model produces much worse geometry (\cref{fig:ablation_on_residual_rgb}), presumable due to the missing of capacity of handling view dependent effects.
It also leads to obvious blurriness in the boundary when rendering from the side view.

A further degenerated case is to use a RGB image and depth for novel-view rendering, and please check our supp for more comparison.

\begin{figure}[t]
    \centering
    \includegraphics[width=1.0\linewidth]{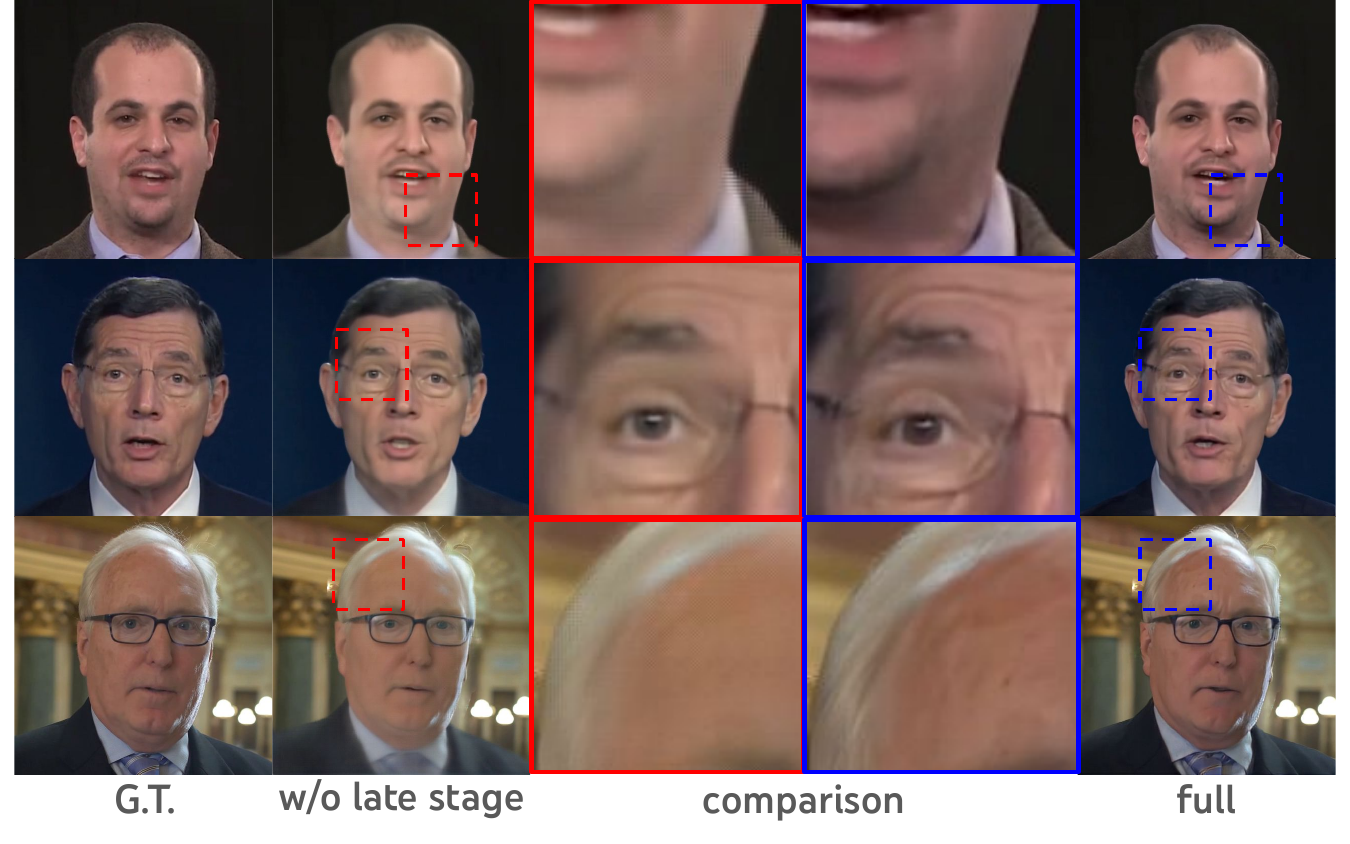}
    \vspace{-2em}
    \caption{\textbf{Effect of Late-stage noise sampling.} We show a comparison between our model trained with and without the Late-stage noise sampling. Training with regular noise sampling leads to excessive blurriness and lost of details.}
    \vspace{-1.7em}
\label{fig:ablation_on_special_noise}
\end{figure}

\noindent\textbf{Late-stage noise sampling.}
Late-stage noise sampling is also important to learn sharp images since that significantly reduces the negative impact from the imperfect pseudo ground truth.
To verify this, we train a model with full noise sampling, and show the quantitative and qualitative comparison in \cref{tab:self-reenact} and \cref{fig:ablation_on_special_noise}.
Without large noise sampling, the diffusion model still produces reasonable structure but missing details, e.g. wrinkles, mold, hair strand.

%% file: sec/5_conclusion.tex
\section{Conclusion}
We present a method to learn a 3D-aware video diffusion model, which generates photorealistic talking head multiplane images given an identity reference image and a sequence of tracking signals.
We show such a model can be learned directly from large scale monocular videos without multi-view datasets.
With appropriate learning strategies, both 3D information and high quality image generation can be learned concurrently.
Extensive experiments show that, our method generates not only talking head video with quality comparable and even better than existing arts but also 3D rendering efficiently that support spatial viewing experience, like in VR headset.
As limitations, the generated talking head video does not support rendering from excessive camera viewpoint changes, which is a common issue of the MPI representation.
Extending our framework with other efficient 3D representations or enhancing training with multi-view dataset could be a promising direction to improve both the image and 3D quality.

%% file: sec/X_suppl.tex
\clearpage
\renewcommand\thesection{\Alph{section}}
\renewcommand\thetable{\Alph{table}}
\renewcommand\thefigure{\Alph{figure}}

\setcounter{section}{0}
\setcounter{figure}{0}
\setcounter{table}{0}
\setcounter{page}{1}
\maketitlesupplementary

In this supplementary material, we provide more information about implementation details (\cref{sec:impl}) and additional results, including more qualitative and quantitative comparisons with other methods (\cref{sec:sota}, \cref{sec:more_compare}, \cref{sec:explain_comparisions}), evaluations on out-of-distribution data (\cref{sec:ood_data}), evaluations on pseudo ground truth images (\cref{sec:pseudo_gt}), 
evaluations on side view renderings (\cref{sec:side_view}) and comparison with ``2D diffusion + depth'' (\cref{sec:rgb_depth}). Additionally, we add further discussions on limitations (\cref{sec:more_limitations}).
We also provide a project page containing video results, as well as stereo videos that can be viewed using a smart phone with Google Cardboard.

\section{Implementation Details}
\label{sec:impl}

\paragraph{Network Architecture}
We build our model architecture based on Space-Time U-Net~\cite{bar2024lumiere}. We first downsample $512\times512$ images into $128\times128$ feature maps using one convolution layer. Then the feature map is downsampled into $64\times64$ by another convolution layer. In the color branch, both downsample blocks and upsample blocks consist of 2 Convolution-based Inflation blocks for resolution 64 and 3 Convolution-based Inflation blocks for resolution 32 and 16. We also add spatial and temporal self-attention layers inside each block with resolution 16.
In the geometry branch, each resolution only has one Convolution-based Inflation block. We apply 
cross-attention layers between the source image features and every block with 32 resolution in the color branch. 
The output feature map is firstly upsampled by pixel shuffle from $64\times64$ into $128\times128$ and processed by a convolution layer. In the color branch, the result $128\times128$ feature map is then upsampled into outputs of $512\times512$ by another pixel shuffle and convolution layer. In the geometry branch, $128\times128$ feature map is firstly upsampled into $256\times256$ and then upsampled into $512\times512$ by pixel shuffle and convolutions.
The color branch predicts the estimated frontal/residual RGB image of the Multiplane Images (MPIs), while the geometry branch predicts alpha images of all planes. Since the geometry branch does not take reference image as condition, reference image features are passed through zero-convolutions between the color branch and the geometry branch.

\paragraph{Training Details}
In the data preparation stage, the reference and target pair of images are randomly selected from the video following~\cite{ma2024follow, xie2024x}, where the camera poses are obtained through 3DMM fitting.

During the model bootstrapping described in main paper Sec.~3.3, we only train the color branch with loss $\mathcal{L}^{\text{mpi}}_{\theta}$ Eq.~(6).
After bootstrapping, we freeze the color branch except the output layer. Then, we train the geometry branch, the output layer of the color branch, and all the zero convolution layers with both losses $\mathcal{L}^{\text{mpi}}_{\theta}$ Eq.~(6) and $\mathcal{L}^{\text{side}}_{\theta}$ Eq.~(7).
In practice, in each iteration, we randomly select one loss from the two to compute the gradients for updating the model. 
We select the loss $\mathcal{L}^{\text{mpi}}_{\theta}$ (Eq.~(6)) with a probability of 0.8, where MPIs are constructed using target cameras. We select the loss $\mathcal{L}^{\text{side}}_{\theta}$ (Eq.~(7)) with a probability of 0.2, where MPIs are constructed using reference cameras, and target cameras serve as side-view cameras.

During training, the weight of LPIPS loss is set to $0.1$ while the mask loss has a weight of $0.01$. The mask loss is formulated as L2 loss since the matting mask has soft boundary. The depth smoothing loss is calculated by a Laplacian kernel and has a weight of $0.01$. The disparity loss has a weight of $0.001$. The drop rate of first frame is $0.7$ while the drop rate of the reference portrait is $0.1$.

\paragraph{Rendering Details}
\label{para:rendering_details}
Following ~\cite{zhao2022generative}, we choose Multiplane Images (MPIs) as our scene representation. We set near and far planes of MPIs adaptively based on the distance $r$ from the MPI frontal camera to the head joint of the parametric model. We place the near plane at $r-0.15$ while the far plane at $r+0.05$. 

During inference, we handle camera via two methods: 1) Rendering the generated MPIs into explicit cameras $\{\phi^{\text{side}}\}$; 2) Rasterizing the 3DMM UV coordinate maps into freely selected MPI frontal cameras as the diffusion controlling signals $\{\mathbf{C}\}$. In our experiments, all the side view renderings, stereo renderings and rendering speed measurements are conducted through the first method. When we render novel views through the first method, novel camera views must stay within a specifiv range of the MPI frontal camera to avoid pose deviations affecting rendering.

\input{tabs/more_compare}

\paragraph{Long Video Classifiers-Free-Guidance (CFG)}

During inference, we assign two different CFG scales to first frame condition and reference portrait condition following Brooks \etal~\cite{brooks2023instructpix2pix}. The guidance scale of the reference portrait is $1.5$. The guidance scale of the first frame is linearly decreased from $1.0$ to $0.0$ from the beginning to the middle of the video clip while the remaining 16 frames in the clip shares a scale of $0.0$.

\begin{figure}[tp]
    \centering
    \includegraphics[width=\linewidth]{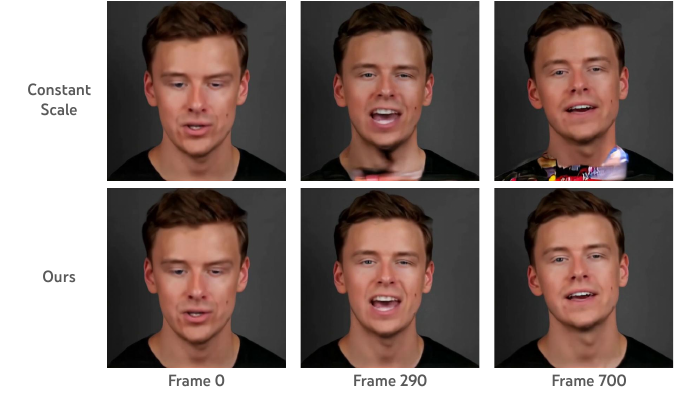}
    \caption{\textbf{Ablation on first frame's classifier-free guidance.} We compare the generated results using a constant classifier-free guidance scale for the first frame versus our scheduled guidance scale.}
    \vspace{-1.5em}
\label{fig:supp_ablation_cfg}
\end{figure}

We show the ablation on first frame's classifier-free guidance in Fig.\ref{fig:supp_ablation_cfg}. We compare our scheduled scale with constant guidance scale of $1.5$. As show in Fig.\ref{fig:supp_ablation_cfg}, artifacts become more pronounced as additional frames are generated. However, using our scheduled scale helps prevent the accumulation of errors.

\section{Additional Qualitative Results}
\label{sec:sota}

\begin{figure*}[tp]
    \centering
    \includegraphics[width=\linewidth]{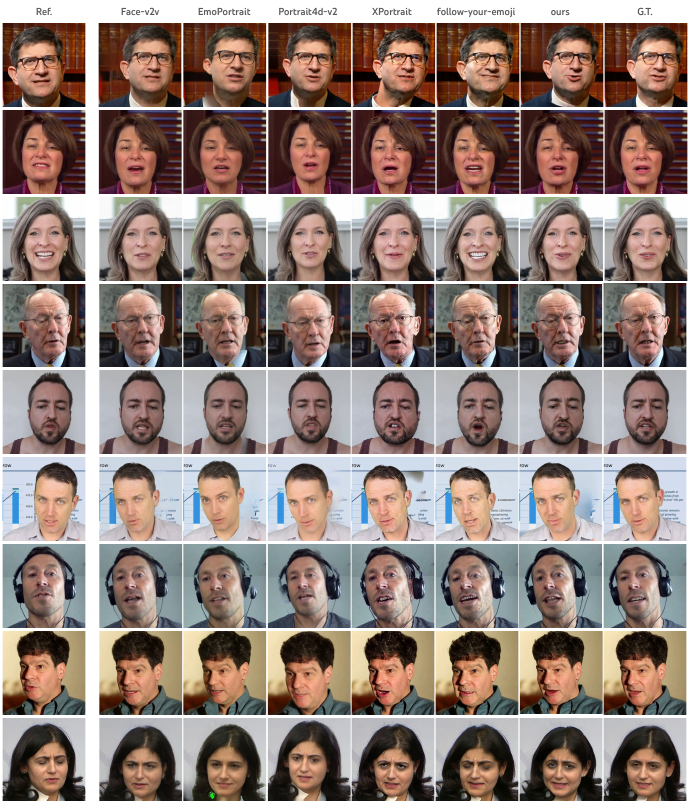}
    \caption{\textbf{Talking head results.} We show more results from previous work and our method.}
\label{fig:supp_comparison}
\end{figure*}

We show more qualitative results on HDTF and Talkinghead1kh datasets in Fig.\ref{fig:supp_comparison}. Face-V2V~\cite{wang2021one} shows good consistency against the driving signal and the reference portrait, while losing details due to its low output resolution. EmoPortrait~\cite{drobyshev2024emoportraits} gives sharp renderings but suffers from identity shifts. Portrait4D-v2~\cite{deng2024portrait4dv2} shows better identity consistency, but losses subtle facial movements. XPortrait~\cite{xie2024x} and Follow-your-emoji~\cite{ma2024follow} generates sharp and high quality frames but suffers from expression and head pose misalignment against the driving signal. Our method generates sharp results that faithfully following driving signals while keeping the identity of generated avatar aligned with the reference portrait. For more comparisons on video quality, please refer to our project page in supplementary material.

\begin{figure}[tp]
    \centering
    \includegraphics[width=0.7\linewidth]{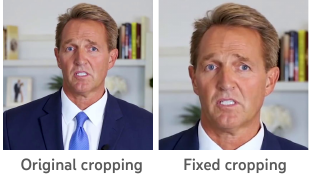}
    \vspace{-1.0em}
    \caption{\textbf{Fixed cropping.} We preprocess some portraits with a tighter cropping around head.}
    \vspace{-0.5em}
\label{fig:supp_cropping}
\end{figure}

Before evaluation, we fix the cropping of all the evaluation data with a tight cropping around head as show in Fig.\ref{fig:supp_cropping}.

\section{Additional Quantitative Results}
\label{sec:more_compare}

We show more qualitative results on VFHQ~\cite{xie2022vfhq} and self-collected dataset in Tab.\ref{tab:more_compare}. For VFHQ dataset, we sample the first 100 frames from all evaluation identities. The self-collected dataset has 50 identities, while each has 32 frames. Consistent with the other two datasets in the main paper, ours achieves the best video quality (FVD) while being competitive in other image-based metrics on VFHQ dataset. On self-collected dataset, ours achieves the best L1, LPIPS and FVD scores. The cross-dataset evaluations following~\cite{chang2023magicpose} reinforces the generation quality and generalizability of our model.

\section{Further Explanations on Comparisons}
\label{sec:explain_comparisions}

\input{tabs/psnr_and_ssim}

In Tab.1 in main paper, we report L1 scores instead of PSNR and SSIM, as these latter metrics are more suitable for pixel-wise alignment in neural rendering tasks, while LPIPS better aligns with perceptual image quality and tolerates some misalignment~\cite{trevithick2023}. As shown in Tab. \ref{tab:psnr_and_ssim}, PSNR and SSIM closely correlate with L1 scores reported in Tab.1, where our method performs worse than non-generative methods like Face-V2V but outperforms others. Due to the generative nature of diffusion models, our method has slightly lower L1 scores but achieves better LPIPS, as it tolerates misalignments. As shown in both Tab.1 and Tab.\ref{tab:psnr_and_ssim}, while sharing network capacity for geometry and learning without 3D data results in a slightly lower FID compared to purely 2D video methods (e.g., Follow-your-emoji, X-Portrait), we still achieve competitive results across all image quality metrics and the best FVD score, which evaluates both image and temporal quality.

\begin{table}
  \centering
    \begin{tabular}{@{}c cc}
        \toprule
        Method & FVD $\downarrow$ & FID $\downarrow$ \\ \hline
        Face-V2V~\cite{wang2021one} & 134.92 & 23.29 \\ 
        EmoPortrait~\cite{drobyshev2024emoportraits} & 197.87 & 34.54 \\ 
        Portrait4d-v2~\cite{deng2024portrait4dv2} & 185.42 & 28.86 \\
        Follow-your-emoji~\cite{ma2024follow} & 154.14 & 20.88 \\
        X-Portrait~\cite{xie2024x} & 199.27 & 21.03 \\ \hline
        Ours & 107.90 & 18.00 \\ 

        \hline
        \bottomrule
    \end{tabular}\
    \caption{\textbf{Evaluation of rendering quality on cross-identity reenactment.}}
  \label{tab:supp_cross}
\end{table}

\begin{figure*}[tp]
    \centering
    \includegraphics[width=0.92\linewidth]{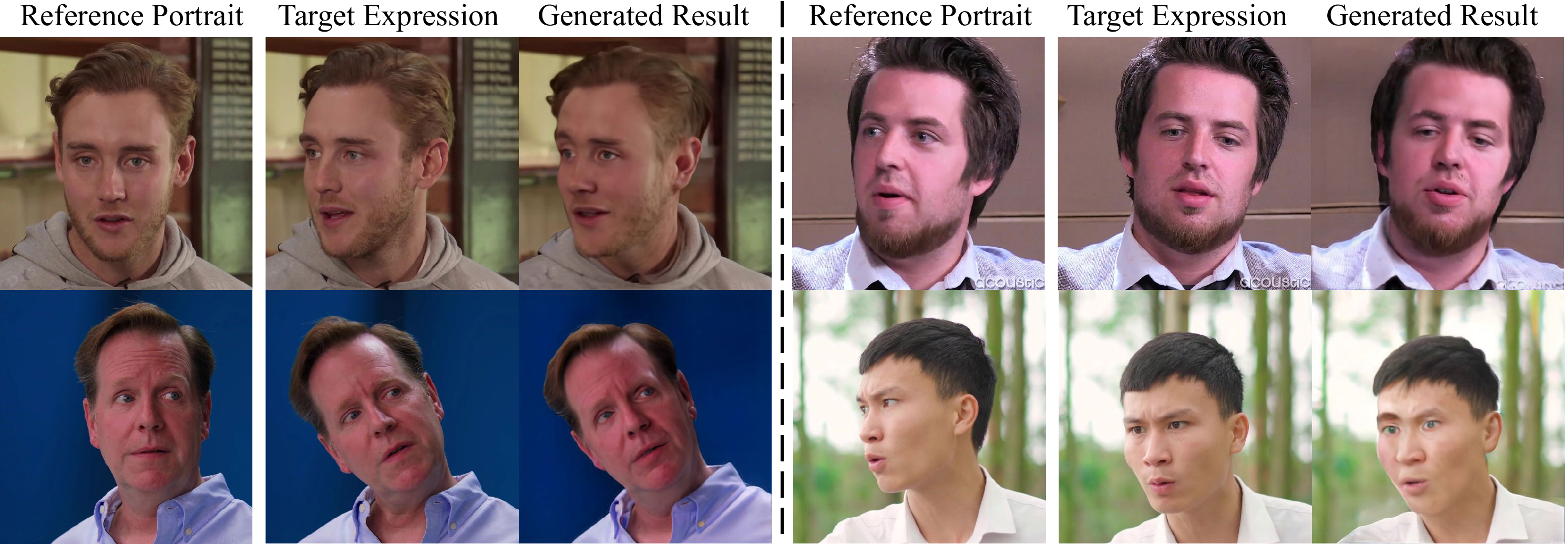}
    \caption{\textbf{Evaluation on large head pose movements.} We show generated results in the MPI frontal camera where the target head poses deviate largely from reference portrait's head poses. In the last row, generated result doesn't align with ground truth since the reference portrait doesn't contain any information about the other side of the character.}
\label{fig:large_head_pose}
\end{figure*}

\begin{figure*}[tp]
    \centering
    \includegraphics[width=0.92\linewidth]{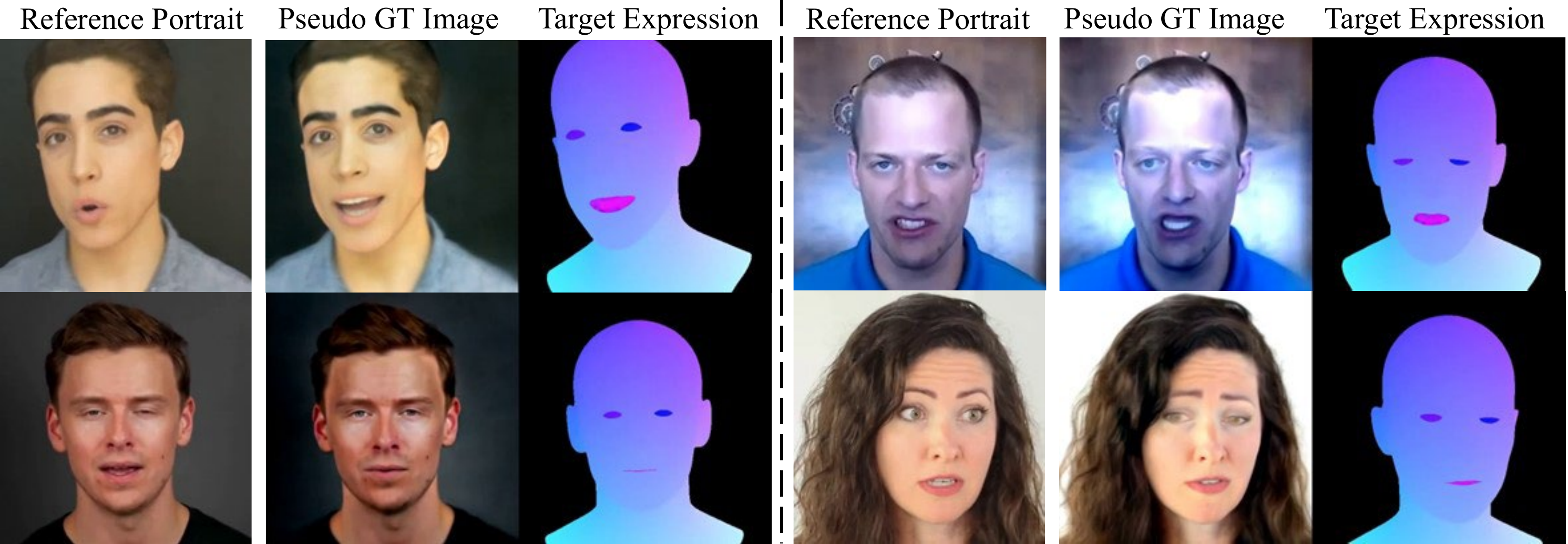}
    \caption{\textbf{Evaluation on pseudo ground truth images.} We show pseudo ground truth images used in the \textbf{Reference-Target Alternating Training}. Pseudo images are only used when large noise is sampled. Though there exist drifting in the color tone, the global structure of generated result aligns with target expressions. }
\label{fig:pseudo_gt_supp}
\end{figure*}

\section{Evaluation on Out-of-distribution Data}
\label{sec:ood_data}

\begin{figure}[tp]
    \centering
    \includegraphics[width=\linewidth]{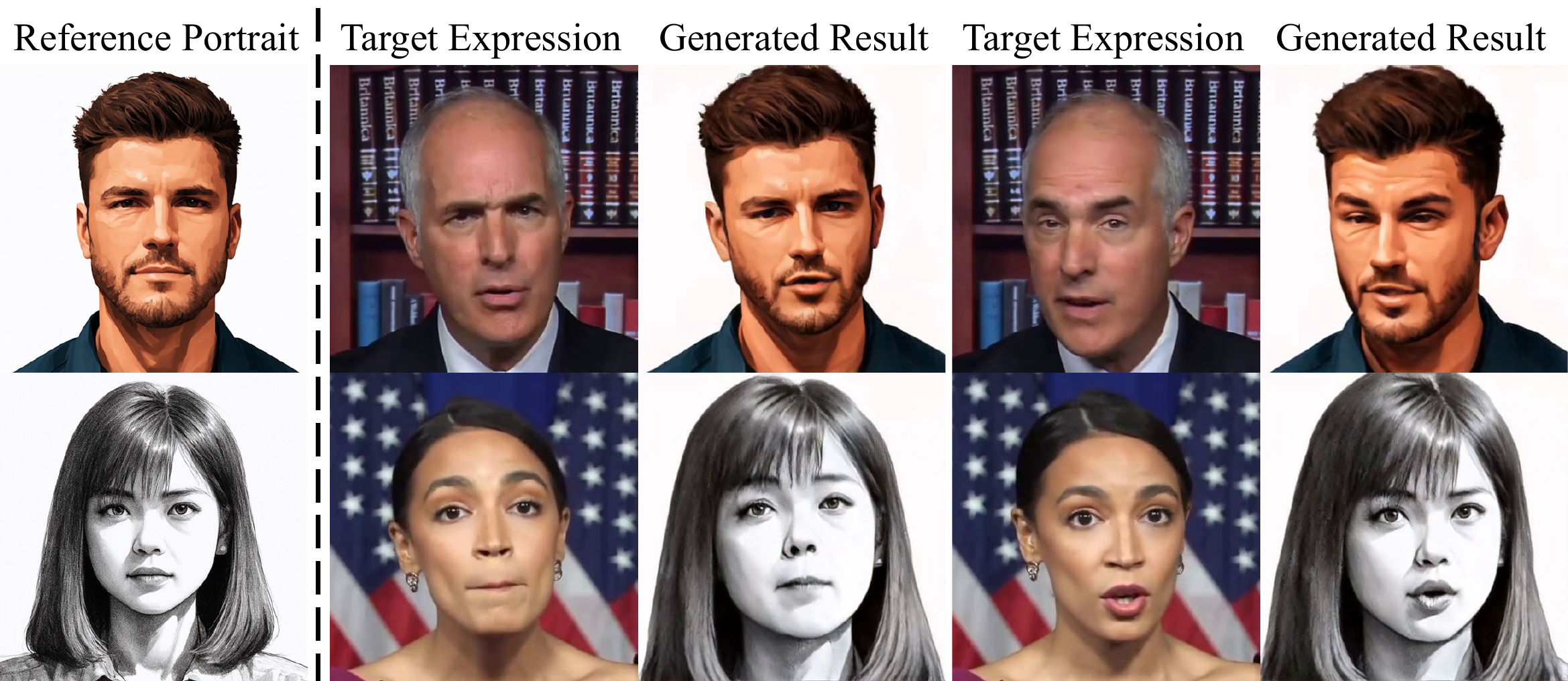}
    \caption{\textbf{Evaluation on synthetic images.} We show generated results when taking synthetic images as reference images. Though trained on a real-world distribution, our model is able to generalize to synthetic images even with a huge gap between the color distribution.}
\label{fig:test_synthetic}
\end{figure}

We further test our model on out-of-distribution data. Despite trained on real world talking-head videos, our model still generalizes to stylized portraits generated by Stable-Diffusion 3~\cite{esser2024scaling}. Results are shown in Fig.\ref{fig:test_synthetic}.

We also evaluate our model on large head pose movements between reference portraits and target expressions through the second method mentioned in \textbf{Rendering Details} of Sec.\ref{para:rendering_details}. As shown in Fig.\ref{fig:large_head_pose}, our model generalize to large head pose movements. However, in some extreme cases, e.g. reference portrait only includes side view of the head, our model may struggle to generate aligned results with ground truth. 

\section{Evaluation on Pseudo Ground Truth Images}
\label{sec:pseudo_gt}

In Fig.\ref{fig:pseudo_gt_supp}, we show some examples of pseudo ground truth images mentioned in Reference-Target Alternating Training section. The generated pseudo ground truth images aligns with target expressions. Note that the pseudo images are only used in training when noise is large, so they don’t need to be sharp and clear as long as the global structure is correct.

\section{Evaluation on Side Views}
\label{sec:side_view}

\begin{figure*}[tp]
    \centering
    \includegraphics[width=\linewidth]{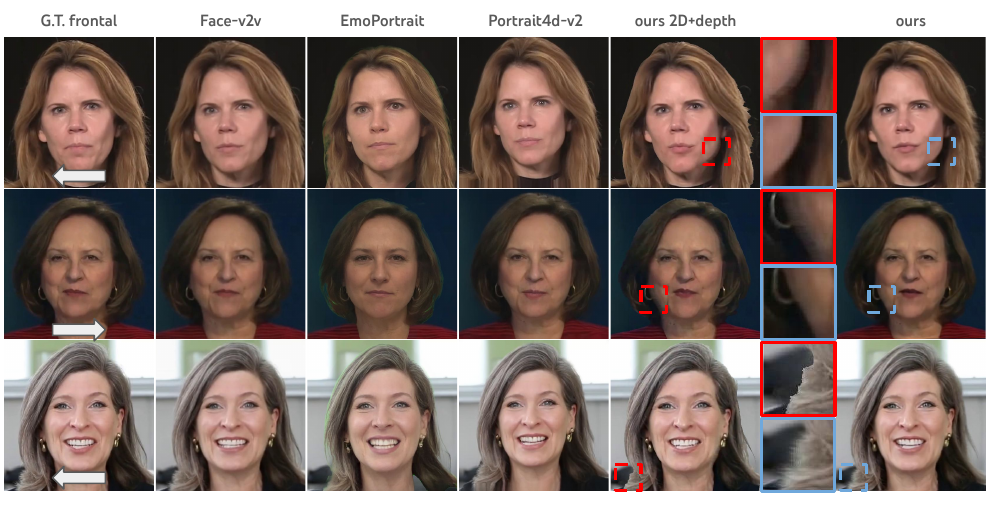}
    \caption{\textbf{Side view rendering comparisons.} We annotate rotation directions by arrows in the ground truth frontal images.}
\label{fig:supp_side_compare}
\end{figure*}

We evaluate our method on side view renderings with HDTF dataset. We select Face-V2V~\cite{wang2021one}, EmoPortrait~\cite{drobyshev2024emoportraits}, Portrait4D-v2~\cite{deng2024portrait4dv2} as baselines since they have 3D representations or implicit 3D feature volumes that support explicit camera control. On the other hand, the diffusion-based baselines~\cite{ma2024follow,xie2024x} only take in images or eyes and mouth landmarks, thus do not enable explicit camera control. 
More specifically, we render several frames by applying a random horizontal viewpoint change (rotate the camera around a fixed look-at point) uniformly sampled from $\pm5^\circ$, and compute the FID score.
All the evaluations are performed in a self-reenactment way. 
The evaluation results are shown in Tab.\ref{tab:supp_side_view}, where our method outperforms all the baselines on FID score.
For more visualizations of our method's side view renderings, please refer to the project page in supplementary material.

\begin{table}
  \centering
    \begin{tabular}{@{}c c}
        \toprule
        Method & FID $\downarrow$ \\ \hline
        View Point & Random within $\pm 5^{\circ}$ \\ \hline
        Face-V2V~\cite{wang2021one} & 22.27 \\ 
        EmoPortrait~\cite{drobyshev2024emoportraits} & 27.71 \\ 
        Portrait4d-v2~\cite{deng2024portrait4dv2} & 27.83 \\
        Ours 2D + Depth~\cite{depth_anything_v2}  & 19.72 \\ \hline
        Ours  & 18.12 \\ 

        \hline
        \bottomrule
    \end{tabular}\
    \caption{\textbf{Evaluation of side view rendering quality}. Our method outperforms all the baseline on rendering quality.}
  \label{tab:supp_side_view}
\end{table}

\section{Compare With 2D Diffusion and Depth Prediction}
\label{sec:rgb_depth}

We also build up a baseline with 
the 2D video diffusion variant of our model and a monocular depth predictor ~\cite{depth_anything_v2} to lift the images into 3D. To be specific, we firstly run depth predictor on all the generated frames. The predicted monocular depth is then aligned with depth of 3DMM mesh by calculating scale and shift. Since the 3DMM mesh doesn't have hair and cloth geometry, we sample depth values on detected facial landmarks as inputs for alignment. 
Before rendering, we first project aligned depth into 3D then connect 3D points of adjacent pixels to create triangles. We also use foreground matting mask to filter out background vertices. During rendering, we use the same background image used by our method.

We compare our method with this baseline and show FID comparisons in Tab.\ref{tab:supp_side_view}. We also show qualitative comparisons in Fig.\ref{fig:supp_side_compare}. Using monocular depth estimator to lift 2D diffusion to 3D suffers from blurry results in occluded regions. Moreover, this baseline also fails to render detailed geometry such as hair as shown in the last row of Fig.\ref{fig:supp_side_compare}.

\section{Additional Discussions on Limitations}
\label{sec:more_limitations}

As discussed in the main paper, our model mainly focuses on realistic foreground human rendering, while background is generated by an inpainting network following~\cite{drobyshev2024emoportraits}. Thus, the generated results do not include animation in the background. To achieve more realistic generated results for background, future works could train an auxiliary network to animate the inpainted background. 
Furthermore, as shown in Fig.\ref{fig:large_head_pose}, large self-occlusions in the reference portrait may lead to degraded results in the unseen facial region. Future work could focus on introducing facial symmetry priors to mitigate this issue.

%% file: tabs/more_compare.tex
\begin{table*}
  \centering
  \small
    \begin{tabular}{c cccc cccc}
    \toprule
    \multirow{2}{*}{Method} & \multicolumn{4}{c}{VFHQ} & \multicolumn{4}{c}{Self-Collected Dataset} \\ 
    \cline{2-5} \cline{6-9} & L1 $\downarrow$ & LPIPS $\downarrow$ & FID $\downarrow$ & FVD $\downarrow$ & L1 $\downarrow$ & LPIPS $\downarrow$ & FID $\downarrow$ & FVD $\downarrow$ \\ \hline
    Face-V2V          & \cellcolor[HTML]{FFA0A0}0.051 & 0.290 & 71.58 & 226.42 & \cellcolor[HTML]{FFD0D0}0.050 & 0.208 & 47.13 & \cellcolor[HTML]{FDEDEC}405.34 \\
    EmoPortrait       & 0.064 & \cellcolor[HTML]{FDEDEC}0.251 & 48.96 & 292.01 & 0.071 & 0.236 & 52.85 & 506.70 \\
    Portrait4d-v2     & - & - & 54.26 & 329.58 & - & - & 54.46 & 553.42 \\
    Follow-your-emoji & \cellcolor[HTML]{FDEDEC}0.057 & \cellcolor[HTML]{FFA0A0}0.198 & \cellcolor[HTML]{FFD0D0}32.40 & \cellcolor[HTML]{FDEDEC}214.29 & \cellcolor[HTML]{FDEDEC}0.060 & \cellcolor[HTML]{FFD0D0}0.181 & \cellcolor[HTML]{FFD0D0}34.94 & \cellcolor[HTML]{FFD0D0}364.36 \\
    X-Portrait        & 0.061 & \cellcolor[HTML]{FFD0D0}0.204 & \cellcolor[HTML]{FFA0A0}26.22 & \cellcolor[HTML]{FFD0D0}211.62 & 0.062 & \cellcolor[HTML]{FDEDEC}0.185 & \cellcolor[HTML]{FFA0A0}32.77 & 499.67 \\
    Ours              & \cellcolor[HTML]{FFD0D0}0.054 & \cellcolor[HTML]{FFD0D0}0.204 & \cellcolor[HTML]{FDEDEC}33.10 & \cellcolor[HTML]{FFA0A0}201.41 & \cellcolor[HTML]{FFA0A0}0.048 & \cellcolor[HTML]{FFA0A0}0.174 & \cellcolor[HTML]{FDEDEC}36.98 & \cellcolor[HTML]{FFA0A0}342.78 \\
    \bottomrule
    
    \end{tabular}
    
    \caption{\textbf{More comparisons on VFHQ and self-collected dataset.} We further compare our method with baselines on both VFHQ and self-collected dataset. Our method achieves comparable image quality and the best video quality (FVD), which is consistent with quantitative comparison in the main paper.}
    \vspace{-1.5em}
  \label{tab:more_compare}
\end{table*}

%% file: tabs/psnr_and_ssim.tex
\begin{table}
  \centering
  \small
    \begin{tabular}{c cccc}
    \toprule
    \multirow{2}{*}{Method} & \multicolumn{2}{c}{HDTF} & \multicolumn{2}{c}{Talkinghead1kh} \\ 
            \cline{2-3} \cline{4-5} & PSNR $\uparrow$ & SSIM $\uparrow$ & PSNR $\uparrow$ & SSIM $\uparrow$ \\ \hline
    Face-V2V          & \cellcolor[HTML]{FFA0A0}27.00 & \cellcolor[HTML]{FFA0A0}0.865 & \cellcolor[HTML]{FFA0A0}24.17 & \cellcolor[HTML]{FFA0A0}0.823 \\
    EmoPortrait       & 22.35 & 0.794 & 19.59 & 0.731 \\
    Follow-your-emoji & 24.16 & 0.811 & 21.50 & 0.759 \\
    X-Portrait        & \cellcolor[HTML]{FDEDEC}24.54 & \cellcolor[HTML]{FDEDEC}0.820 & \cellcolor[HTML]{FDEDEC}21.82 & \cellcolor[HTML]{FDEDEC}0.767 \\
    Ours              & \cellcolor[HTML]{FFD0D0}24.83 & \cellcolor[HTML]{FFD0D0}0.833 & \cellcolor[HTML]{FFD0D0}22.43 & \cellcolor[HTML]{FFD0D0}0.777 \\
    \bottomrule
    
    \end{tabular}
    \caption{\textbf{PSNR and SSIM on HDTF and Talkinghead1kh datasets.} PSNR and SSIM rankings align with the L1 score rankings in Table 1 of the main paper.}
    \vspace{-1.0em}
  \label{tab:psnr_and_ssim}
\end{table}